\documentclass[sigconf]{acmart}

\AtBeginDocument{%
  }

\copyrightyear{2025}
\acmYear{2025}
\setcopyright{rightsretained}
\acmConference[UIST Adjunct '25]{The 38th Annual ACM Symposium on User Interface Software and Technology}{September 28-October 1, 2025}{Busan, Republic of Korea}
\acmBooktitle{The 38th Annual ACM Symposium on User Interface Software and Technology (UIST Adjunct '25), September 28-October 1, 2025, Busan, Republic of Korea}\acmDOI{10.1145/3746058.3758426}
\acmISBN{979-8-4007-2036-9/2025/09}

\begin{document}

\title[Set the Stage]{Set the Stage: Enabling Storytelling with Multiple Robots through Roleplaying Metaphors}

\author{Tyrone Justin {Sta. Maria}}
\orcid{0000-0002-6826-7890} 
\affiliation{%
  \institution{De La Salle University}
 \city{Manila}
  \country{Philippines}}
\email{tyrone_stamaria@dlsu.edu.ph}

\author{Faith {Griffin}}
\orcid{0009-0004-3978-767X} 
\affiliation{%
  \institution{De La Salle University}
 \city{Manila}
  \country{Philippines}}
\email{faith_griffin@dlsu.edu.ph}

\author{Jordan Aiko {Deja}}
\orcid{0001-9341-6088}
 \affiliation{%
  \institution{De La Salle University}
  \city{Manila}
  \country{Philippines}}
\email{jordan.deja@dlsu.edu.ph}

\renewcommand{\shortauthors}{Sta. Maria, Griffin \& Deja}

\begin{abstract}

Gestures are an expressive input modality for controlling multiple robots, but their use is often limited by rigid mappings and recognition constraints. To move beyond these limitations, we propose roleplaying metaphors as a scaffold for designing richer interactions. By introducing three roles: Director, Puppeteer, and Wizard, we demonstrate how narrative framing can guide the creation of diverse gesture sets and interaction styles. These roles enable a variety of scenarios, showing how roleplay can unlock new possibilities for multi-robot systems. Our approach emphasizes creativity, expressiveness, and intuitiveness as key elements for future human-robot interaction design.

\end{abstract}


\begin{teaserfigure}
\centering
  \includegraphics[width=1\textwidth]{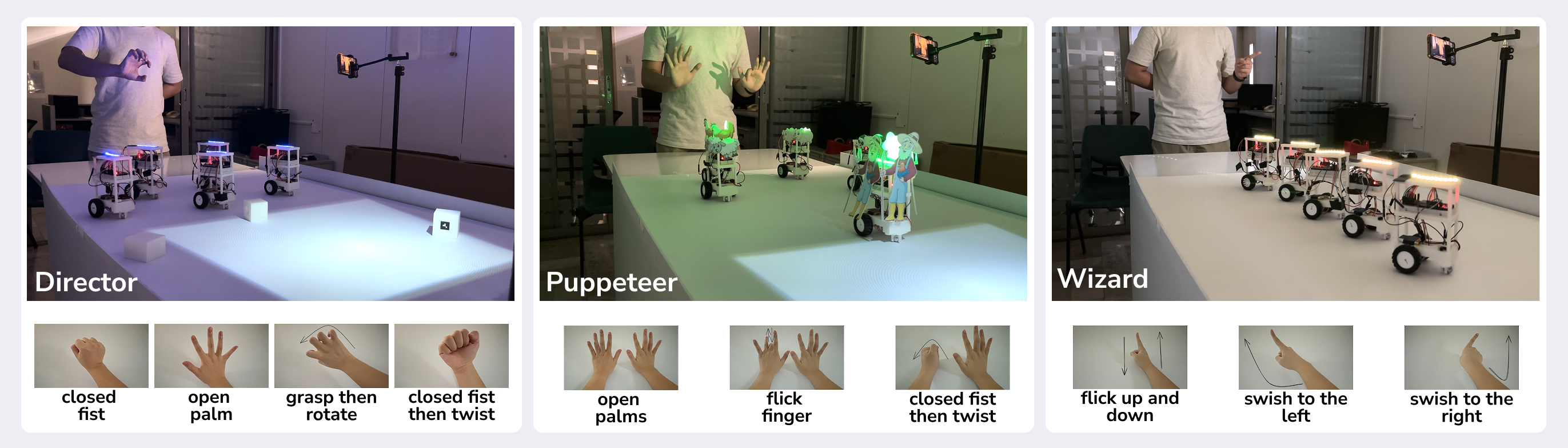}
  \caption{The \textit{Set the Stage} interaction space. The top row illustrates the three roleplaying metaphors—\textit{Director}, \textit{Puppeteer}, and \textit{Wizard}—each offering a distinct mode of control. The bottom row shows a gesture vocabulary of hand and finger movements that can be combined and repurposed to expand the interaction space and support a variety of application scenarios.}
  \label{fig:teaser}
\end{teaserfigure}


\maketitle

\section{Introduction}
\par Interacting with multiple robots is becoming an increasingly relevant challenge as robots are gradually integrated into homes, studios, classrooms, and performance spaces~\cite{bauer2008human, suzuki2022augmented, goodrich2008human, sheridan2016human}. In these contexts, gesture-based interaction presents itself as a powerful and expressive modality. However, the design of gesture interfaces for multi-robot systems remains limited~\cite{abioye2024adaptive, kaduk2024one}. Most systems rely on static mappings between gestures and robot actions~\cite{le2016zooids,wang2024push,kim2020user, ichihashi2024swarm}, often optimized for task efficiency or recognition accuracy rather than user expressiveness or adaptability~\cite{liu2018gesture}. These constraints restrict creativity and make it difficult for users to develop rich, meaningful interactions with robots~\cite{salem2011friendly}.

\par In this work, we present \textit{Set the Stage}, an interaction space that uses roleplaying metaphors~\cite{wingren2024using, ng2024role} to scaffold gesture-based interaction with multiple robots~\cite{brunnmayr2024approaching}. Rather than assigning gestures to isolated commands, we frame interaction through three performative roles: (i) Director, (ii) Puppeteer, and (iii) Wizard, each offering a distinct gestural vocabulary and control logic. These metaphors are drawn from familiar narrative structures~\cite{rond2019improv, collins2021does}: a director choreographs, a puppeteer animates, and a wizard casts. Framing gestures in this way helps users conceptualize robot control not just as command issuance, but as expressive, semi-structured performance~\cite{alves2021collection}.


\par By grounding gesture design in metaphor and narrative structure~\cite{dennler2023design, la2024train}, our approach enables more flexible, imaginative, and improvisational interactions even when working with a limited set of gesture-to-action mappings. A small set of roles can already unlock a wide range of application scenarios, illustrating how roleplay can serve as a creative scaffold for future gesture design~\cite{sandoval2022human, elgarf2021once}. \textit{Set the Stage} thus contributes a design perspective that shifts gesture-based HRI beyond technical recognition constraints and into a space of play, narrative thinking, and expressive interaction with particular focus to contexts such as teaching, storytelling, prototyping, and even live performance.

\section{Set the Stage Interaction Space}

\begin{figure}
    \centering
    \includegraphics[width=1\linewidth]{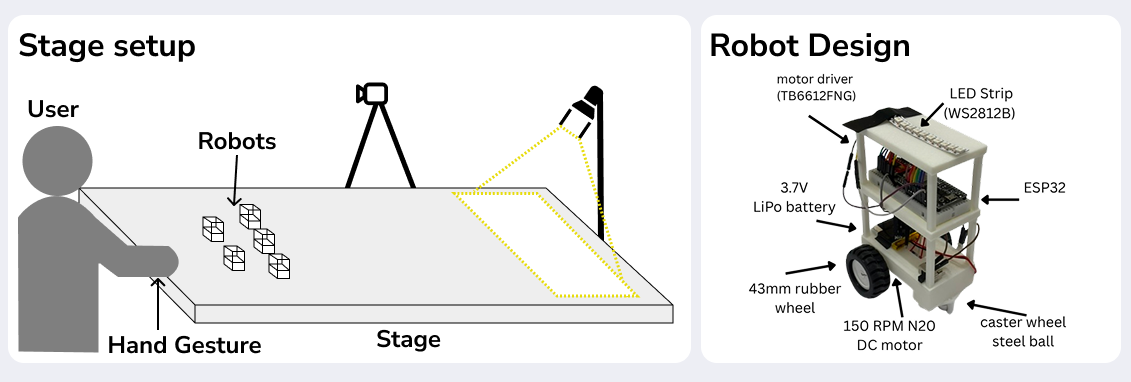}
    \caption{Left: The interaction stage includes robots, and a spotlight, with the user standing front-facing and a gesture-tracking camera positioned at the side. Right: The custom-built robot features dual N20 DC motors, rubber wheels, an ESP32 microcontroller, and an LED strip, all powered by a 3.7V LiPo battery and enclosed in a compact 3D-printed chassis.}
    \label{fig:space-design}
\end{figure}
\par Building on the conceptual framing introduced in Section~1, we developed \textit{Set the Stage} as an interaction space that brings roleplaying metaphors into embodied actions with multiple robots (see ~\autoref{fig:space-design}). Set the Stage uses three narrative roles namely Director, Puppeteer, and Wizard to organize and re-purpose a limited set of gestures into meaningful, context-driven robot behaviors. These metaphors offer users distinct modes of interaction, each framing gesture use in a different way and enabling varied applications despite working with the same foundational gesture set.\\

\par \noindent \textbf{Director:} As a Director, the user treats the robot group as an ensemble cast on stage. Inspired by how theater directors use hand signals to guide actors, this role frames gestures as direct, high-level commands for collective movement. For example, pushing an open palm forward drives the robot group ahead, while pulling a fist back initiates reverse motion. A rotating wrist gesture, preceded by a grasp, causes the group to pivot left or right (see ~\autoref{fig:teaser} top left). Though simple, these movements are choreographed with how the Director navigates the group through scenes, avoids obstacles, and positions the swarm into spotlights, creating a structured and narrative-driven spatial flow. 

\par Extending this metaphor, the Director can re-purpose the very same gesture set to craft tightly synchronized, dance-like sequences. By chaining simple cues such as forward, backward, rotate into timed patterns, the ensemble transitions from mere locomotion to expressive group choreography. These scripted motions transform the stage into a dance stage, where consistency of timing and spacing amplifies narrative impact while still leveraging the Director’s concise, high-level command vocabulary. \\


\par \noindent \textbf{Puppeteer:} The Puppeteer engages at a more granular level, controlling individual robots through finger-based gestures. Each robot is linked to a finger, turning the hand into a living marionette controller. Flicking a finger forward moves its corresponding robot, while collective movements (e.g., forming a fist and rotating the wrist) trigger synchronized responses. In our implementation, these robots take on the roles of characters from \textit{Old MacDonald Had a Farm} (see ~\autoref{fig:teaser} top middle), with the Puppeteer guiding them into position in a sequenced narrative. This role showcases how a simple gesture vocabulary can be reused in precise, character-driven contexts.


\par Given that individual fingers can already control specific robots (or groups), what if the marionette metaphor extended beyond the cast? We could imagine using the same fine-grained gestures not only to animate robots, but also to interact with elements of the environment such as props, lighting, or backdrops. For example, hand-based gestures of the Director allow switching control modes between characters and stage elements, while finger gestures could sequence or ``cue'' both robot and prop actions into a coordinated storyline. This opens the door for richer, narrative-driven interaction where the user becomes both performer and stage manager.\\


\par \noindent \textbf{Wizard:} Finally, the Wizard role re-imagines gesture as magical casting. The user’s index finger becomes a wand, performing swift directional swishes and flicks to trigger visual and behavioral effects. A vertical flick toggles the robots' lights, while horizontal swishes combine rotation and movement to choreograph group actions (see ~\autoref{fig:teaser} top right). Here, gestures are not just commands. Instead, they are spells that animate the robots into dramatic light and motion effects. In our prototype, this role enables the user to simulate a lightning effect, using rapid gesture sequences to create dynamic, audiovisual patterns.


\par What if we combined the wand-like swishes and flicks of the Wizard with the precise, finger-based control of the Puppeteer? Fingers could guide individual robots to specific positions on stage, while swishes synchronize their LED lights as they move or perform. This combination could enable choreographed sequences illustrating how constellations align, how each fairy blesses Sleeping Beauty in a familiar animated tale, or how drones are coordinated during a nighttime search-and-rescue mission. The gestures remain simple: swish, flick, poke, repeat, just like your fairy godmother might have done, but behind them lies a rich language for expressive, multi-robot storytelling.

\section{Conclusion and Future Work}

\par In this work, we introduced \textit{Set the Stage}, a gesture-based interaction paradigm for controlling multiple robots through roleplaying metaphors: Director, Puppeteer, and Wizard. Each role reframes a limited gesture set into distinct interaction styles, enabling a broader range of expressive, narrative-driven behaviors. Across all roles, gestures are intentionally reused and repurposed, highlighting how metaphor extends their meaning and application beyond their limited gesture mappings. This layering of narrative structure onto physical input acts as a creative scaffold, allowing users to explore richer interaction possibilities within the same gesture space.

\par While we have yet to conduct user studies, we view this work as a starting point for rethinking how metaphors can guide multi-robot interaction design. Future work includes evaluating usability, expressiveness, and creativity support through controlled studies, and exploring new role metaphors that further expand opportunities for improvisational and storytelling-based interaction.

\bibliographystyle{ACM-Reference-Format}
\bibliography{main}

\appendix









\end{document}